\documentclass[letterpaper, 10 pt, journal, twoside]{ieeetran}  

\IEEEoverridecommandlockouts                              

\usepackage[
    style=ieee,
    doi=false,
    isbn=false,
    url=false,
    eprint=false,
    backend=bibtex,
    natbib=true,
    minnames=1,
    maxcitenames=1
    ]{biblatex}

\usepackage[shortcuts]{extdash}
\usepackage{graphicx}
\usepackage{mathtools}
\usepackage{dcolumn}
\usepackage{gensymb}
\usepackage{perl_acronyms}
\usepackage{xcolor}
\usepackage{amsfonts}
\usepackage{booktabs}
\usepackage{siunitx}
\usepackage[inline]{enumitem}
\usepackage{array}
\usepackage{flushend}
\newcolumntype{P}[1]{>{\centering\arraybackslash}p{#1}}

\DeclarePairedDelimiter\abs{\lvert}{\rvert}%

\title{SilhoNet: An RGB Method for 6D Object Pose Estimation}

\author{Gideon Billings$^{1}$ and Matthew Johnson-Roberson$^{1}$%
\thanks{Manuscript received: Feb, 24, 2019; Revised May, 24, 2019; Accepted July, 9, 2019.}
\thanks{This paper was recommended for publication by Editor Cesar Cadena upon evaluation of the Associate Editor and Reviewers' comments.
This work was supported by the NASA award NNX16AL08G} 
\thanks{$^{1}$Gideon Billings and Matthew Johnson-Roberson are with Department of Naval Architecture and Marine Engineering, University of Michigan, 2600 Draper Dr. Ann Arbor, MI 48109, USA
        {\tt\footnotesize gidobot@umich.edu}}%
\thanks{Digital Object Identifier (DOI): see top of this page.}
}

\begin{document}

\IEEEpubid{\begin{minipage}[t]{\textwidth}\ \\[25pt]
        \centering\scriptsize{Copyright \copyright 2019 IEEE Personal use is permitted. For any other purposes, permission must be obtained from the IEEE by emailing pubs-permissions@ieee.org}
\end{minipage}}

\maketitle
\IEEEpubidadjcol

\begin{abstract}
\label{sec:abstract}

Autonomous robot manipulation involves estimating the translation and orientation of the object to be manipulated as a 6-degree-of-freedom (6D) pose. Methods using RGB-D data have shown great success in solving this problem. However, there are situations where cost constraints or the working environment may limit the use of RGB-D sensors. When limited to monocular camera data only, the problem of object pose estimation is very challenging. In this work, we introduce a novel method called SilhoNet that predicts 6D object pose from monocular images. We use a Convolutional Neural Network (CNN) pipeline that takes in \ac{ROI} proposals to simultaneously predict an intermediate silhouette representation for objects with an associated occlusion mask and a 3D translation vector. The 3D orientation is then regressed from the predicted silhouettes. We show that our method achieves better overall performance on the YCB-Video dataset than two state-of-the art networks for 6D pose estimation from monocular image input.
\end{abstract}


\section{INTRODUCTION}
\label{sec:introduction}

\IEEEPARstart{R}{obots} are revolutionizing the way technology enhances our lives. From helping people with disabilities perform various tasks around their house to autonomously collecting data in humanly inaccessible environments, robots are being applied across a spectrum of exciting and impactful domains. Many of these applications require the robot to grasp and manipulate an object in some way (e.g., opening a door by a handle, or picking up an object from the seafloor), but this poses a challenging problem. Specifically, the robot must interpret sensory information of the scene to localize the object. Beyond robot manipulation, there are also applications, such as augmented reality, which require accurate localization of an object in an image.

Previous methods for object pose estimation largely depend on RGB-D data about the 3D working environment~\cite{bohg_data-driven_2014, papazov_rigid_2012, miyazaki_object_2017, azizi_geometric_2017}. However, there are cases where such depth information is not readily available. Some examples include systems that operate outdoors where common depth sensors like the Kinect do not work well, embedded systems where space and cost may limit the size and number of sensors, and underwater vehicles where the variable and scattering properties of the water column result in noisy and sparse depth information. In these scenarios, methods that operate on monocular camera data are needed. When the sensor modality is limited to monocular images, estimating the pose of an object in a natural setting is a challenging problem due to variability in scene illumination, the variety of object shapes and textures, and occlusions caused by scene clutter.

Recently, there has been progress in state-of-the-art methods for monocular image pose estimation on difficult datasets, where the scenes are cluttered and objects are often heavily occluded~\cite{cao_real-time_2016,rad_bb8:_2017, xiang_posecnn:_2017, tekin_real-time_2017, kehl_ssd-6d:_2017, li_unified_2018, li_deepim:_2018}. The presented work improves on the performance of these recent methods to deliver a novel deep learning based method for 6D object pose estimation on monocular images. Unlike prior methods, we explicitly incorporate prior knowledge of the 3D object appearance into the network architecture, and we make use of an intermediate silhouette based object viewpoint representation to improve on orientation prediction accuracy. Further, this method provides occlusion information about the object, which can be used to determine which parts of an object model are visible in the scene. Knowing how the target object is occluded in the monocular image can be important for certain applications, such as augmented reality, where it is desirable to project over only the visible portion of an object.

In this paper, we present the following contributions:
\begin{enumerate*}
    \item SilhoNet, a novel RGB-based deep learning method to estimate pose and occlusion in cluttered scenes;
    \item The use of an intermediate silhouette representation to facilitate learning a model on synthetic data to predict 6D object pose on real data, effectively bridging the sim-to-real domain shift~\cite{2018arXiv180307721C};
    \item A method to determine which parts of an object model are visually unoccluded, using the projection of inferred silhouettes, in novel scenes;
    \item An evaluation on the visually challenging YCB-Video dataset~\cite{xiang_posecnn:_2017} where the proposed approach outperforms two state-of-the-art RGB method.
\end{enumerate*}

The rest of this paper is organized in the following sections: section II discusses related work; section III presents our method with an overview of our CNN design for 6D pose estimation and occlusion mask prediction; section IV presents the experimental results; and section V concludes the paper.
\vspace{-6pt}
\section{RELATED WORK}
\label{sec:related}

\begin{figure*}[t!]
    \centering
    \includegraphics[width=\textwidth]{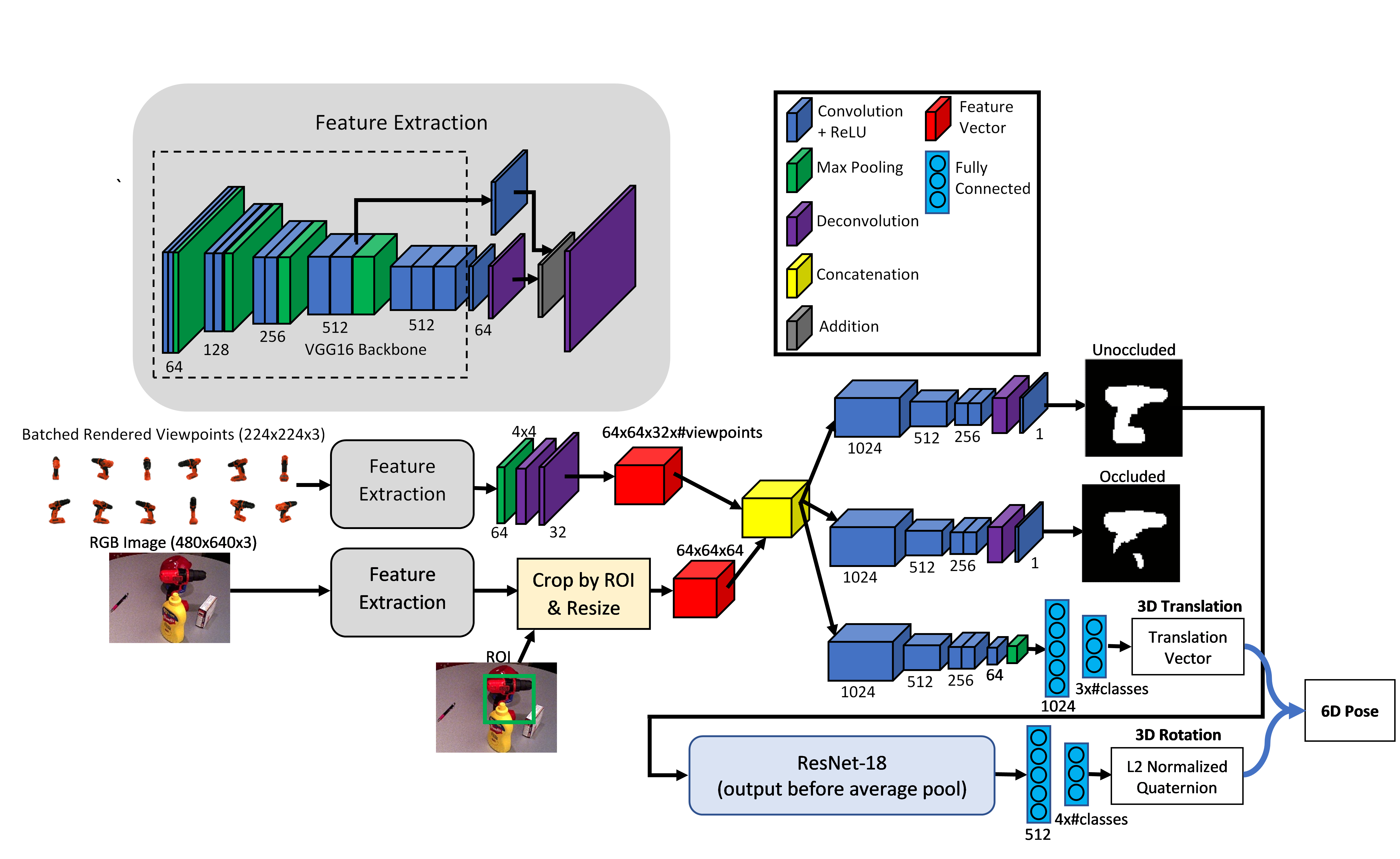}
            \vspace{-27pt}
    \caption{Overview of the SilhoNet pipeline for silhouette prediction and 6D object pose estimation. The 3D translation is predicted in parallel with the silhouettes. The predicted unoccluded silhouette is fed into a second stage network to predict the 3D rotation vector.}
    \label{fig:architecture}
    \vspace{-20pt}
\end{figure*}

Extensive research has focused on 6D object pose estimation using RGB\=/D data. Several works rely on feature- and shape-based template matching to locate the object in the image and coarsely estimate the pose~\cite{hinterstoisser_model_2013, 6751365, cao_real-time_2016}. This is often followed by a refinement step using the Iterative Closest Point (ICP) algorithm with the 3D object model and a depth map of the scene~\cite{hinterstoisser_model_2013}. While these methods are computationally efficient, their performance often degrades in cluttered environments. Other methods have exploited point cloud data to match 3D features and fit the object models into the scene~\cite{5540108, 10.1007/978-3-319-46487-9_51, 10.1007/978-3-319-46487-9_51}. While point cloud based methods achieve state-of-the-art performance, they can be very computationally expensive. Recent works have demonstrated the power of machine learning for object detection and pose estimation using RGB\=/D data. \citet{7139363} used a CNN pretrained on ImageNet to extract features from an RGB image and a colorized depth map. They learned a series of Support Vector Machines (SVM) on top of these extracted features to predict the object category and a single axis rotation about a planar surface normal. In~\cite{10.1007/978-3-319-10605-2_35}, they trained a decision forest to regress every pixel from an RGB\=/D image to an object class and a coordinate position on the object model. Other work has used a CNN to map the pose of an object in an observed RGB\=/D image to a rendered pose of the model through an energy function~\cite{krull}. The minimization of the energy function gives the object pose. \citet{Michel2017GlobalHG} trained a Conditional Random Field (CRF) to output a number of pose-hypotheses from a dense pixel wise object coordinate prediction map computed by a random forest. A variant of ICP was used to derive the final pose estimate. While these learning-based methods are powerful, efficient, and give state-of-the-art results, they rely on RGB\=/D data to estimate the object pose.

There are several recent works extending deep learning methods to the problem of 6D object pose estimation using RGB data only. \citet{rad_bb8:_2017, tekin_real-time_2017} used a CNN to predict 2D projections of the 3D object bounding box corners in the image, followed by a PnP algorithm to find the correspondences between the 2D and 3D coordinates and compute the object pose. \citet{xiang_posecnn:_2017} proposed a multistage, multibranch network with a Hugh Voting scheme to directly regress the 6D object pose as a 3D translation and a unit quaternion orientation. \citet{kehl_ssd-6d:_2017} predicted 2D bounding box detections with a pool of candidate 6D poses for each box. After a pose refinement step, they choose the best candidate pose for each box. \citet{li_unified_2018} used an end-to-end CNN framework to predict discretely binned rotation and translation values with corrective delta offsets. They proposed a novel method for infusing the class prior into the learning process to improve the network performance for multi-class prediction. \citet{li_deepim:_2018} proposed a deep-learning-based iterative matching algorithm for RGB based pose refinement, which achieves performance close to methods that use depth information with ICP and can be applied as post refinement to any RGB based method. These RGB-based pose estimation methods demonstrate competitive performance against state-of-the-art approaches that rely on depth data. Our work extends these recent advancements in monocular pose estimation by combining the power of deep learning with prior knowledge of the object model to estimate pose from silhouette predictions. Also, our method provides information about how the object is visually occluded in the form of occlusion masks, which can be projected onto the object model, given the predicted 3D orientation.

\section{METHOD}
\label{sec:method}

We introduce a novel method that operates on monocular color images to estimate the 6D object pose. The 3D orientation is predicted from an intermediate unoccluded silhouette representation. The method also predicts an occlusion mask which can be used to determine which parts of the object model are visible in the image. The method operates in two stages, first predicting an intermediate silhouette representation and occlusion mask of an object along with a vector describing the 3D translation and then regressing the 3D orientation quaternion from the predicted silhouette. The following sections describe our method in detail.

\subsection{Overview of the Network Pipeline}

Figure~\ref{fig:architecture} presents an overview of the network pipeline. The input to the network is an RGB image with \ac{ROI} proposals for detected objects and the associated class labels. The first stage uses a VGG16~\cite{VGG16} backbone with deconvolution layers at the end to produce a feature map from the RGB input image. This feature extraction network is the same as used in PoseCNN~\cite{xiang_posecnn:_2017}. Extracted features from the input image are concatenated with features from a set of rendered object viewpoints and then passed through three network branches, two of which have identical structure to predict a full unoccluded silhouette and an occlusion mask. The third branch predicts a 3D vector encoding the object center in pixel coordinates and the range of the object center from the camera. The second stage of the network passes the predicted silhouette through a ResNet-18~\cite{ResNet18} architecture with two fully connected layers at the end to output an L2-normalized quaternion, representing the 3D orientation.

\subsubsection{Predicted ROIs}

We trained an off-the-shelf Faster-RCNN implementation from Tensorpack~\cite{Tensorpack} on the YCB-video dataset~\cite{xiang_posecnn:_2017} to predict \ac{ROI} proposals. This network was trained across two Titan V GPUs for 3,180,000 iterations on the training image set with the default parameters and without any synthetic data augmentation. The \ac{ROI} proposals are provided as input to the network after the feature extraction stage, where they are used to crop the corresponding region out of the input image feature map. The cropped feature map is then resized to a width and height of 64x64 by either scaling down the feature map or using bi-linear interpolation to scale it up.

\subsubsection{Rendered Model Viewpoints}

We were able to boost the silhouette prediction performance by generating a set of synthetic pre-rendered viewpoints associated with the detected object class as an additional input to the first stage of the network. For each class, we rendered a set of 12 viewpoints from the object model, each with dimension 224x224. These viewpoints were generated using Phong shading at azimuth intervals from 0\degree\ to 300\degree\ with elevation angles of -30\degree\ and 30\degree. As the intermediate goal is silhouette prediction, these synthetic renders are able to capture the shape and silhouette of real objects, in different orientations, despite the typical domain shift in the visual appearance of simulated objects~\cite{2018arXiv180307721C}.

All the viewpoints for the detected object class are passed through the feature extraction stage and then resized to 64x64 with channel dimension 32 by passing them through a max-pooling layer with width 4 and stride 4, followed by two deconvolution layers that each increase the feature map size by 4. In our implementation, we extracted the feature maps of the rendered viewponts on-the-fly for each object detection. However, to reduce computation time, these extracted feature maps can be precomputed and stored offline. These rendered viewpoint feature maps were provided to the network by stacking them on the channel dimension and then concatenating with the cropped and resized input image feature map~(Fig.\ref{fig:architecture}).

\subsubsection{Silhouette Prediction}

The first stage of the network predicts an intermediate silhouette representation of the object as a 64x64 dimensional binary mask.  This silhouette represents the full unoccluded visual hull of the object as though it were rendered with the same 3D orientation but centered in the frame. The size of the silhouette in the frame is invariant to the scale of the object in the image and is determined by a fixed distance of the object from the camera at which the silhouette appears to be rendered. This distance is chosen for each object so that the silhouette just fits within the frame for any 3D orientation. Given the smallest field of view of the camera $A$ determined by the minimum of the width and height of the image sensor, the 3D extent of the object as the width, height and depth $(w,h,d)$, we calculate the render distance $r$ as
\vspace{-4pt}
\begin{equation}
    \vspace{-4pt}
    {\displaystyle r = 1.05\frac{\sqrt{w^2 + h^2 + d^2}}{2\tan(A/2)}.}
\end{equation}
\noindent
This stage of the network also has a parallel branch that outputs a similar silhouette, with only the unoccluded parts of the object visible. We refer to this occluded output as the `occlusion mask'.

The first part of the network is a VGG16 feature extractor~\cite{VGG16}, which generates feature maps at 1/2, 1/4, 1/8, and 1/16 scale. The 1/8 and 1/16 scale feature maps both have an output channel dimension of 512. The channel dimension for both is reduced to 64 using two convolution layers, after which the 1/16 scale map is upscaled by a factor of 2 using deconvolution and then summed with the 1/8 scale map. The summed map is upscaled by a factor of 8 using a second deconvolution to get a final feature map of the same dimension as the input image with a feature channel width of 64~(Fig.\ref{fig:architecture}).

After the input image is passed through the feature extractor, the input \ac{ROI} proposal for the detected object is used to crop out the corresponding area of the resulting feature map and resize it to 64x64. This feature map is concatenated with the rendered viewpoint feature maps, resulting in a single feature vector matrix with size 64x64x448.

The feature vector matrix is fed into two identical network branches, one of which outputs the silhouette prediction and the other outputs the occlusion mask. Each branch is composed of 4 convolution layers, each with a filter width, channel dimension, and stride of (2, 1024, 1), (2, 512, 2), (3, 256, 1), and (3, 256, 1) respectively, followed by a deconvolution layer with filter width, channel dimension, and stride of (2, 256, 2). The output of the deconvolution layer is fed into a dimension reducing convolution filter with a single channel output shape of 64x64. A sigmoid activation function is applied at the output to produce a probability map.

\subsubsection{3D Translation Regression}

The 3D translation is predicted as a three dimensional vector, encoding the object center location in pixel coordinates and range from the camera center in meters. Other region proposal based pose estimation methods~\cite{xiang_posecnn:_2017, do2018deep} regress the Z coordinate directly from the \ac{ROI}. However, this suffers from ambiguities. If an object at a given range is shifted along the arc formed by the circle with the camera center as the focus, the Z coordinate will change while the object appearance in the shifted \ac{ROI} will be unchanged. This ambiguity is especially prevalent in wide field of view cameras. By predicting the object range rather than directly regressing the Z coordinate, our method does not suffer from ambiguities and can recover the Z coordinate with good accuracy. Given the camera focal length $f$, the pixel coordinates of the object center $(px,py)$ with respect to the image center, and the range $r$ of the object center form the camera center, similar triangles can be used to show that the 3D object translation, $(X,Y,Z)$, can be recovered as

\vspace{-10pt}
\begin{equation}
    {\displaystyle Z = \frac{rf}{\sqrt{px^2 + py^2 + f^2}},}
\end{equation}
\begin{equation}
    {\displaystyle X = Z*px/f, \quad Y = Z*py/f.}
    \vspace{-4pt}
\end{equation}
\noindent
The pixel coordinates of the object center are predicted with respect to the \ac{ROI} box as an offset from the lower box edge bounds normalized by the box dimensions and passed through a sigmoid function. Given a \ac{ROI} with width $w$, height $h$, lower x and y coordinate bounds $(bx,by)$, the coordinates of the image principal point $(cx,cy)$ and the predicted normalized output from the network $(nx,ny)$, the object center pixel coordinates $(px,py)$ are recovered as
\vspace{-4pt}
\begin{equation}
    {\displaystyle rx = -\log(1/nx - 1), \quad ry = -\log(1/ny - 1),}
    \vspace{-4pt}
\end{equation}
\begin{equation}
    {\displaystyle px = bx + rx*w - cx, \quad px = by + ry*h - cy.}
    \vspace{-4pt}
\end{equation}
\noindent
Note that only the pixel coordinates of the object center are offset by the principal point in these equations. While other methods limit the prediction of the object center to lie within the \ac{ROI}~\cite{xiang_posecnn:_2017} or treat the \ac{ROI} center as the coordinates of the object center~\cite{do2018deep}, if the object is not completely in the image frame, the center may not lie within the \ac{ROI}, and because \ac{ROI} predictions are imperfect, the object center rarely lies at the \ac{ROI} center. Our formulation for predicting the object center does not constrain the point to lie within the \ac{ROI} and is robust to imperfect \ac{ROI} proposals.

The translation prediction branch is identical to the silhouette prediction branches, except the deconvolution layer is replaced with a 5th convolution layer with filter width, channel dimension, and stride of (2, 64, 2) followed by max pooling. The output is fed into a fully connected layer of dimension 1024 followed by a fully connected layer of dimension 3x(\# classes), where each class has a separate output vector. The predicted vector for the class of the detected object is extracted from the output, and the first two entries are normalized with a sigmoid activation~(Fig.\ref{fig:architecture}).

\subsubsection{3D Orientation Regression}

We use a quaternion representation for the 3D orientation, which can represent arbitrary 3D rotations in continuous space as a unit vector of length 4. The quaternion representation is especially attractive, as it does not suffer from gimbal lock like the Euler angle representation. Predicting orientation from a \ac{ROI} gives rise to visual ambiguities, as the true object orientation varies depending on the location within the image from which the \ac{ROI} is extracted. To address these ambiguities, the network predicts the apparent orientation as though the \ac{ROI} were extracted from the center of the image. Given the predicted object translation, the true orientation is recovered by applying a pitch, $\delta\theta$, and roll, $\delta\phi$, adjustment to the predicted orientation. These adjustments are calculated as
\vspace{-6pt}
\begin{equation}
    {\displaystyle \delta\theta = \arctan(X/Z), \quad \delta\phi = -\arctan(Y/Z),}
    \vspace{-6pt}
\end{equation}

The second stage of the network takes in the predicted silhouette probability maps, thresholded at some value into binary masks, and outputs a quaternion prediction for the object orientation. This stage of the network is composed of a ResNet-18~\cite{ResNet18} backbone, with the layers from the average pooling and below replaced with two fully connected layers. The last fully connected layer has output dimension 4x(\# classes), where each class has a separate output vector. The predicted vector for the class of the detected object is extracted from the output and normalized using an L2-norm~(Fig.\ref{fig:architecture}).

Because the silhouette representation of objects is featureless, this method treats symmetries in object shape as equivalent symmetries in the 3D orientation space. In many robotic manipulation scenarios, this is a valid assumption. For example, a tool such as a screwdriver that may not be symmetric in RGB feature space is symmetric in shape and equivalently symmetric in grasp space. However, it is a future goal of this work to extend the 3D orientation estimation to account for non-symmetries in feature space.

By regressing the 3D orientation from an intermediate silhouette representation, we were able to train this stage of the network using only synthetically rendered silhouette data. In our results, we show that the network generalized well to predicting pose on real data, showing that this intermediate representation as an effective way to bridge the domain shift between real and synthetic data.

\subsubsection{Occlusion Prediction}

Given the predicted apparent 3D orientation of the object, we can project the predicted occlusion mask onto the object model to determine which portions of the model are visible in the scene. Mathematically, this can be accomplished by taking every vertex $v$ of the object model and projecting it onto the occlusion mask. We construct a transform matrix $T$ with a z translation component equal to the render distance $r$ for the corresponding object class and the x and y translation components set to 0. The rotation sub-matrix is formed from the predicted apparent orientation. Using the following equation, each vertex of the object model can be projected onto the occlusion mask, which is scaled up to fit the minimum dimension of the input image,

\vspace{-4pt}
\begin{equation}
    {\displaystyle \gamma = KTv}
    \vspace{-4pt}
\end{equation}
\noindent
where $K$ is the camera intrinsic matrix, $v$ is the 3D homogeneous coordinates of the vertex in the object frame, and $\gamma$ is the homogeneous pixel coordinates of the projected vertex on the scaled occlusion mask. Not accounting for object self occlusions, those vertices which lie on the visible portion of the occlusion mask are predicted to be visible in the image.

\subsection{Dataset}

We evaluated our method on the YCB-video dataset~\cite{xiang_posecnn:_2017}, which consists of 92 video sequences composed of 133,827 frames, containing a total of 21 objects, appearing in different arrangements with varying levels of occlusion. Twelve of the video sequences were withheld from the training set for validation and testing. In the silhouette space, the objects in this dataset are characterized by five different types of symmetry: non-symmetric, symmetric about a plane, symmetric about two perpendicular planes, symmetric about an axis, symmetric about an axis and a plane. We applied a rotation correction to the coordinate frame of all objects that exhibit any form of symmetry so that each axis or plane of symmetry aligns with a coordinate axis. Ground truth quaternions were generated from the labeled object poses such that only one unique quaternion is associated with every viewpoint that produces the same visual hull. Having a consistent quaternion label for all matching silhouette viewpoints enabled the pose prediction network to be trained effectively for all types of object symmetries using a very simple distance loss function.

Supplementing the real image data in the YCB-video dataset are 80,000 synthetically rendered images, with all of the 21 objects appearing in various combinations and random poses over a transparent background. We supplement the training data by randomly sampling images from the COCO-2017 dataset~\cite{COCO2017} and applying them as background to these synthetic images at training time.

\subsection{Network Training}

All networks were trained with the Adam optimizer on either a Titan V or Titan X GPU. The VGG16 backbone was initialized with ImageNet pre-trained weights, and the silhouette prediction network without the translation branch was trained using cross entropy loss with a batch size of 6 for 325,000 iterations. We trained the network with ground truth \ac{ROI}s and tested against both ground truth \ac{ROI}s and predicted \ac{ROI}s from a Faster-RCNN network~\cite{Tensorpack} trained on the YCB-video dataset. The translation prediction branch was then added, and all network weights not part of this branch were frozen. The translation branch was trained for 230,000 iterations using an l2 loss. All network weights were then unfrozen and the entire network was finetuned for 208,000 iterations.

The orientation regression network was trained using the following log distance function between the predicted and ground truth quaternions
\vspace{-6pt}
\begin{equation}
    QLoss(\widetilde{q},q) = log(\epsilon + 1 - \abs{\widetilde{q} \cdot q}),
    \vspace{-4pt}
\end{equation}

where q is the ground truth quaternion, $\widetilde{q}$ is the predicted quaternion, and $\epsilon$ is a small value for stability, in our case $e^{-4}$. The orientation regression network was trained for 380,000 iterations with a batch size of 16, using only perfect ground truth silhouettes for training. Testing was done on the predicted silhouettes from the first stage network.

To reduce overfitting during training of the networks, dropout was applied at a rate of 0.5 before the last deconvolution layer of the feature extraction network, on the fourth convolutional layer of each silhouette prediction branch, and after the max pooling layer of the translation branch. During training of the orientation regression network, dropout was applied at a rate of 0.8 before the first fully connected layer. As a further strategy to reduce overfitting and extend the training data, the hue, saturation, and exposure of the training images were randomly scaled by a factor of up to 1.5

\begin{table}[htbp!]
  \scriptsize
  \centering
  \caption{Mean IoU accuracy for predicted silhouettes}
  \vspace{-6pt}
    \begin{tabular}{p{6.5em}cccc}
    \toprule
    Object & \multicolumn{1}{P{4em}}{Unoccluded GT ROI} & \multicolumn{1}{P{4em}}{Occluded GT ROI} & \multicolumn{1}{P{4em}}{Unoccluded Pred ROI} & \multicolumn{1}{P{4em}}{Occluded Pred ROI} \\
    \midrule
    master\_chef\_can   & 96.75 & 91.08 & 96.84 & 88.42 \\
    cracker\_box        & 92.94 & 82.20 & 90.50 & 68.91 \\
    sugar\_box          & 94.28 & 91.79 & 92.32 & 88.27 \\
    tomato\_soup\_can   & 96.41 & 93.25 & 96.73 & 94.09 \\
    mustard\_bottle     & 95.02 & 94.49 & 94.68 & 94.25 \\
    tuna\_fish\_can     & 95.96 & 93.81 & 96.06 & 93.95 \\
    pudding\_box        & 90.08 & 79.57 & 88.73 & 71.58 \\
    gelatin\_box        & 95.72 & 94.65 & 95.31 & 94.78 \\
    potted\_meat\_can   & 92.53 & 87.11 & 93.77 & 87.18 \\
    banana              & 88.48 & 87.23 & 81.76 & 78.05 \\
    pitcher\_base       & 94.63 & 93.80 & 94.58 & 93.71 \\
    bleach\_cleanser    & 92.48 & 89.64 & 91.74 & 87.95 \\
    bowl                & 79.74 & 67.01 & 82.03 & 76.63 \\
    mug                 & 93.92 & 86.84 & 90.97 & 84.24 \\
    power\_drill        & 86.61 & 85.08 & 78.57 & 73.64 \\
    wood\_block         & 89.30 & 74.92 & 90.72 & 78.84 \\
    scissors            & 52.20 & 65.12 & 61.70 & 65.97 \\
    large\_marker       & 84.37 & 84.15 & 83.96 & 82.65 \\
    large\_clamp        & 84.03 & 79.50 & 85.73 & 80.93 \\
    extra\_large\_clamp & 86.16 & 82.34 & 76.13 & 70.14 \\
    foam\_brick         & 91.00 & 86.17 & 89.99 & 82.78 \\
    \midrule
    ALL   & 89.17 & 85.23 & 88.23 & 82.71 \\
    \bottomrule
    \end{tabular}%
  \label{tab:sih_miou}%
\end{table}%

\section{RESULTS}
\label{sec:results}

\begin{table*}[!ht]
  \scriptsize
  \centering
  \caption{Mean 3D orientation error in degrees. The Sym tag indicates orientation predictions are reduced by geometric symmetries.}
  \vspace{-6pt}
    \begin{tabular}{p{7em}cccc|ccc}
    \toprule
    \multicolumn{1}{c}{} & \multicolumn{4}{c}{RGB} & \multicolumn{2}{|c}{RGB-D} \\
    \cmidrule{2-7}
    Object & \multicolumn{1}{P{4em}}{PoseCNN \cite{xiang_posecnn:_2017}} & \multicolumn{1}{P{4em}}{PoseCNN Sym~\cite{xiang_posecnn:_2017}} &
    \multicolumn{1}{P{4em}}{SilhoNet-GT~ROI} & \multicolumn{1}{P{4em}}{SilhoNet-Pred~ROI} &
    \multicolumn{1}{|P{4.5em}}{PoseCNN +ICP~\cite{xiang_posecnn:_2017}} & \multicolumn{1}{P{5.7em}}{PoseCNN +ICP~Sym~\cite{xiang_posecnn:_2017}} \\
    \midrule
    master\_chef\_can   & 50.71 &  7.57 &   \textbf{1.11} &   1.21 & 51.88 &  1.06 \\
    cracker\_box        & 19.69 & 19.69 &   \textbf{9.53} &  19.86 &  9.51 &  9.23 \\
    sugar\_box          &  \textbf{9.29} &  \textbf{9.29} &  11.50 &  12.28 &  1.06 &  1.06 \\
    tomato\_soup\_can   & 18.23 &  8.40 &   \textbf{1.82} &   1.91 & 31.74 &  1.98 \\
    mustard\_bottle     &  9.94 &  9.59 &   \textbf{5.07} &   5.78 &  2.72 &  2.22 \\
    tuna\_fish\_can     & 32.80 & 12.74 &   1.50 &   \textbf{1.46} & 37.70 &  6.28 \\
    pudding\_box        & \textbf{10.20} & \textbf{10.20} &  18.39 &  20.95 &  2.27 &  2.26 \\
    gelatin\_box        &  \textbf{5.25} &  \textbf{5.25} &   8.48 &  12.52 &  1.03 &  1.03 \\
    potted\_meat\_can   & 28.67 & 19.74 &  10.93 &   \textbf{7.27} & 23.06 & 13.93 \\
    banana              & 15.48 & 15.48 &   \textbf{5.70} &  16.29 & 12.17 & 12.17 \\
    pitcher\_base       & 11.98 & 11.98 &   \textbf{6.61} &   6.64 &  2.55 &  2.55 \\
    bleach\_cleanser    & \textbf{20.85} & \textbf{20.85} &  48.42 &  51.28 & 11.02 & 11.02 \\
    bowl                & 75.53 & 75.53 &  53.95 &  \textbf{49.95} & 55.71 & 55.71 \\
    mug                 & 19.44 & 19.44 &   \textbf{7.02} &  18.14 & 23.11 & 23.11 \\
    power\_drill        &  \textbf{9.91} &  \textbf{9.91} &  10.66 &  30.54 &  1.64 &  1.64 \\
    wood\_block         & 23.63 & 23.63 &  \textbf{23.23} &  25.52 & 15.12 & 15.12 \\
    scissors            & \textbf{43.98} &\textbf{ 43.98} & 154.82 & 155.53 & 30.77 & 30.76 \\
    large\_marker       & 92.44 & 13.59 &  10.72 &  \textbf{10.44} & 84.34 &  3.38 \\
    large\_clamp        & 38.12 & 38.12 &   6.03 &   \textbf{3.54} & 33.99 & 33.99 \\
    extra\_large\_clamp & 34.18 & 34.18 &   \textbf{7.30} &  29.18 & 37.89 & 37.89 \\
    foam\_brick         & 22.67 & 22.67 &  17.36 &  \textbf{13.84} & 18.82 & 18.82 \\
    \midrule
    ALL                 & 27.79 & 17.82 &  \textbf{13.48} &  16.04 & 24.54 & 10.94 \\
    \bottomrule
    \end{tabular}%
  \label{tab:orienation}%
\end{table*}%

\begin{table}[htbp]
  \scriptsize
  \centering
  \caption{Mean 3D translation error in centimeters}
  \vspace{-6pt}
    \begin{tabular}{p{7em}ccc|cc}
    \toprule
    \multicolumn{1}{c}{} & \multicolumn{3}{c}{RGB} & \multicolumn{1}{|c}{RGB-D} \\
    \cmidrule(lr){2-5} 
    Object & \multicolumn{1}{P{4em}}{PoseCNN \cite{xiang_posecnn:_2017}} & \multicolumn{1}{P{4em}}{SilhoNet-GT~ROI} & \multicolumn{1}{P{4em}}{SilhoNet-Pred~ROI} & \multicolumn{1}{|P{4em}}{PoseCNN +ICP~\cite{xiang_posecnn:_2017}} \\
    \midrule
    master\_chef\_can   & 3.29 & 3.14 & \textbf{3.02} & 0.52 \\
    cracker\_box        & 4.02 & \textbf{2.38} & 5.24 & 1.28 \\
    sugar\_box          & 3.06 & \textbf{1.67} & 2.10 & 0.26 \\
    tomato\_soup\_can   & 3.02 & \textbf{2.24} & 2.40 & 0.33 \\
    mustard\_bottle     & 1.72 & \textbf{1.41} & 1.65 & 0.14 \\
    tuna\_fish\_can     & 2.41 & \textbf{1.49} & 1.57 & 0.37 \\
    pudding\_box        & 3.69 & \textbf{1.91} & 7.15 & 0.31 \\
    gelatin\_box        & 2.49 & \textbf{0.79} & 1.09 & 0.19 \\
    potted\_meat\_can   & 3.65 & \textbf{2.74} & 4.30 & 1.06 \\
    banana              & \textbf{2.43} & 2.59 & 4.12 & 0.63 \\
    pitcher\_base       & 4.43 & \textbf{1.29} & 1.31 & 0.14 \\
    bleach\_cleanser    & 4.86 & 3.99 & \textbf{3.60} & 0.49 \\
    bowl                & 5.23 & 4.08 & \textbf{3.30} & 3.73 \\
    mug                 & 4.00 & \textbf{1.43} & 2.61 & 0.97 \\
    power\_drill        & 4.59 & \textbf{3.19} & 6.77 & 0.17 \\
    wood\_block         & 6.34 & \textbf{3.23} & 5.59 & 2.68 \\
    scissors            & 6.40 & \textbf{2.59} & 9.91 & 1.49 \\
    large\_marker       & 3.89 & \textbf{2.31} & 3.24 & 0.89 \\
    large\_clamp        & 9.79 & \textbf{3.51} & 6.27 & 5.25 \\
    extra\_large\_clamp & 8.36 & \textbf{2.12} & 4.86 & 4.19 \\
    foam\_brick         & 2.48 & \textbf{2.31} & 3.98 & 0.48 \\
    \midrule
    ALL                 & 4.16 & \textbf{2.45} & 3.49 & 1.06 \\
    \bottomrule
    \end{tabular}%
  \label{tab:translation}%
\end{table}%

\begin{table}[htbp]
  \scriptsize
  \centering
  \caption{Area under accuracy-threshold curve for 6D pose evaluation using ADD-S metric from \cite{xiang_posecnn:_2017}}
  \vspace{-6pt}
    \begin{tabular}{p{7em}cccccc}
    \toprule
    Object & \multicolumn{1}{P{3em}}{PoseCNN \cite{xiang_posecnn:_2017}} & \multicolumn{1}{P{3.5em}}{SilhoNet-GT~ROI} & \multicolumn{1}{P{3.5em}}{SilhoNet-Pred~ROI} & \multicolumn{1}{P{3em}}{MCN \cite{li_unified_2018}} & \multicolumn{1}{P{3.5em}}{MV5-MCN~\cite{li_unified_2018}} \\
    \midrule
    master\_chef\_can   & 82.6 &  83.6 & 84.0 & 87.8 & \textbf{90.6} \\
    cracker\_box        & 77.2 &  \textbf{88.4} & 73.5 & 64.3 & 72.0 \\
    sugar\_box          & 84.0 &  \textbf{88.8} & 86.6 & 82.4 & 87.4 \\
    tomato\_soup\_can   & 81.7 &  89.4 & 88.7 & 87.9 & \textbf{91.8} \\
    mustard\_bottle     & 91.1 &  91.0 & 89.8 & 92.5 & \textbf{94.3} \\
    tuna\_fish\_can     & 84.0 &  \textbf{89.9} & 89.5 & 84.7 & 89.6 \\
    pudding\_box        & 79.4 &  \textbf{89.1} & 60.1 & 51.0 & 51.7 \\
    gelatin\_box        & 85.7 &  \textbf{94.6} & 92.7 & 86.4 & 88.5 \\
    potted\_meat\_can   & 78.5 &  84.8 & 78.8 & 83.1 & \textbf{90.3} \\
    banana              & 85.9 &  \textbf{88.7} & 80.7 & 79.1 & 85.0 \\
    pitcher\_base       & 76.9 &  \textbf{91.8} & 91.7 & 84.8 & 86.1 \\
    bleach\_cleanser    & 71.5 &  72.0 & 73.6 & 76.0 & \textbf{81.0} \\
    bowl                & 63.5 &  72.5 & 79.6 & 76.1 & \textbf{80.2} \\
    mug                 & 78.1 &  92.1 & 86.8 & 91.4 & \textbf{93.1} \\
    power\_drill        & 72.7 &  \textbf{82.9} & 56.5 & 76.0 & 81.1 \\
    wood\_block         & 61.5 &  \textbf{79.2} & 66.2 & 54.0 & 58.4 \\
    scissors            & 56.6 &  78.3 & 49.1 & 71.6 &\textbf{ 82.7} \\
    large\_marker       & 68.3 &  \textbf{83.1} & 75.0 & 60.1 & 66.3 \\
    large\_clamp        & 55.3 &  \textbf{84.5} & 69.2 & 66.8 & 77.5 \\
    extra\_large\_clamp & 42.8 &  \textbf{88.4} & 72.3 & 61.1 & 68.0 \\
    foam\_brick         & 86.7 &  \textbf{88.4} & 77.9 & 60.9 & 67.7 \\
    \midrule
    ALL                 & 75.3 &  \textbf{85.8} & 79.6 & 75.1 & 80.2 \\
    \bottomrule
    \end{tabular}
  \label{tab:adds}
\end{table}

The following sections present the performance of SilhoNet, tested on the YCB-video dataset~\cite{Calli:2017aa}. Section A presents the accuracy of the silhouette prediction stage, and section B compares the 6D pose estimation performance of SilhoNet against the performance of PoseCNN~\cite{xiang_posecnn:_2017}. We also compare performance against the method in~\cite{li_unified_2018} for RGB input.

\subsection{Silhouette Prediction}

\vspace{4pt}
\begin{figure}
    \centering
    \includegraphics[width=0.8\linewidth]{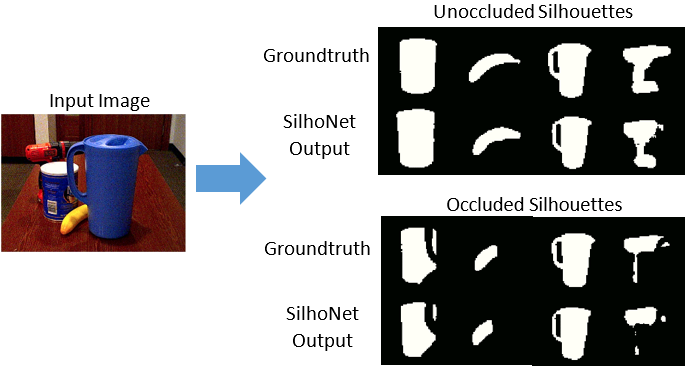}
    \vspace{-10pt}
    \caption{Example prediction of occluded and unoccluded silhouettes from a test image}
    \label{fig:sih_pred}
\end{figure}

We tested the performance of the silhouette prediction stage of SilhoNet with both ground truth \ac{ROI} inputs from the YCB dataset and predicted \ac{ROI} inputs from the FasterRCNN network. Figure~\ref{fig:sih_pred} shows an example of the silhouette predictions for one of the images in the test set. Table~\ref{tab:sih_miou} presents the accuracy for the occluded and unoccluded silhouette predictions, measured as the mean intersection over union (IoU) of the predicted silhouettes with the ground truth silhouettes. Overall, the performance degrades by a few percent when the predicted \ac{ROI}s (Pred ROI) are provided as input rather than the ground truth (GT ROI), but in general, the predictions are robust to the \ac{ROI} input.

\subsection{6D Pose Regression}

\begin{figure}
    \centering
    \includegraphics[width=1.0\linewidth]{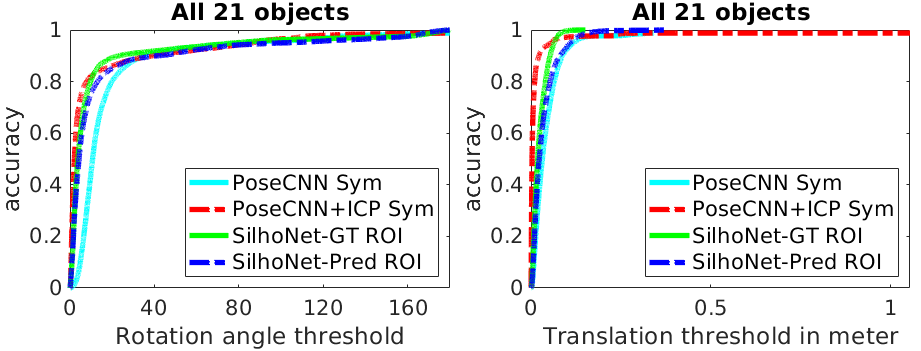}
    \vspace{-20pt}
    \caption{6D pose accuracy curve across all objects in the YCB-video dataset. Accuracy is percentage of errors less than the error threshold. The PoseCNN orientation predictions are reduced by the same geometric symmetries as SilhoNet.}
    \label{fig:mean_acc_curve}
\end{figure}

We compare the accuracy of the 6D pose predictions from SilhoNet against the published results of PoseCNN. We include the performance of PoseCNN with depth based Iterative Closest Point (ICP) refinement as an RGB-D method reference point. To provide greater insight into the model performance, we first analyze the orientation and translation prediction results separately. Because our method predicts orientation in a space reduced by geometric symmetries, we compare against the performance of PoseCNN both before and after reducing the PoseCNN predictions to the same symmetry invariant space. Figure~\ref{fig:mean_acc_curve} shows the accuracy curves for PoseCNN before and after ICP refinement and SilhoNet with YCB ground truth \ac{ROI} input (GT ROI) and FasterRCNN predicted \ac{ROI} input (Pred ROI). SilhoNet shows a visually higher area under the accuracy curve than PoseCNN before ICP refinement. The improvement of SilhoNet in area under the accuracy curve is especially obvious for the rotation angle prediction accuracy, demonstrating the effectiveness of the intermediate silhouette representation for orientation prediction.
Table~\ref{tab:orienation} presents the mean orientation errors for each class across both the PoseCNN and SilhoNet methods. The classes with the worst prediction accuracy for SilhoNet relative to PoseCNN are "bleach\_cleanser" and "scissors". SilhoNet treats both of these objects as non-symmetric in silhouette space, but the shape of both objects is nearly planar symmetric, especially if they are partially occluded, so pose predictions from silhouettes may be easily confused. SilhoNet shows the strongest performance on cylindrical objects like "master\_chef\_can" and "tomato\_soup\_can", which exhibit the highest reduction in orientation space through symmetries. Across every type of geometric symmetry exhibited in the dataset, there are objects where SilhoNet performs significantly better than PoseCNN, demonstrating the general effectiveness of silhouettes as an intermediate representation for object 3D orientation estimation. The orientation prediction accuracy of SilhoNet is reduced when predicted \ac{ROI}s are provided as input, but overall there is still significant improvement over PoseCNN, showing that SilhoNet is robust to the quality of region proposals.

Table~\ref{tab:translation} presents the mean translation errors for each object class. SilhoNet outperforms PoseCNN across most classes before ICP refinement. The translation prediction accuracy of SilhoNet is also reduced when predicted \ac{ROI}s are provided as input, but there is still significant improvement over PoseCNN.

In Table~\ref{tab:adds}, we compare the full 6D pose prediction performance of SilhoNet against PoseCNN (without depth refinement)~\cite{xiang_posecnn:_2017} and another recently proposed RGB based method~\cite{li_unified_2018}. We use the area under the accuracy-threshold curve (ADD-S) metric proposed in \cite{xiang_posecnn:_2017}. The ADD-S metric is particularly suited to SilhoNet, as it is invariant to geometric symmetries. We note that the method MV5-MCN~\cite{li_unified_2018} is a multiview variant of MCN~\cite{li_unified_2018} and requires that each input image is labelled with a camera pose. Typically, labelling camera pose would require some extra sensory input besides a monocular RGB camera in order to disambiguate the scale of motion in a SLAM system. The results in the table show that SilhoNet outperforms PoseCNN and MCN by a large margin with both ground truth and predicted \ac{ROI}s as input. SilhoNet performs better than MV5-MCN with ground truth \ac{ROI}s as input and performs on par with predicted \ac{ROI}s as input. Overall, SilhoNet shows a significant performance improvement over related methods when the input is limited to RGB images only.

\vspace{4pt}

As an ablation study, we performed an experiment to determine the contribution of the rendered viewpoint image priors to the network performance. Table~\ref{tab:model_input} shows the results of this experiment. Note that the network was trained without the translation prediction branch, and ground truth \ac{ROI}s were given as input. When no rendered viewpoints are provided as a prior input, the network performance drops with nearly twice the error in orientation predictions for both shared and class specific output. However, providing more than one rendered viewpoint image as a prior input does not significantly affect the network performance. This result motivates future investigation into how the network incorporates the rendered viewpoint inputs into the learned network structure.

\begin{table}[htbp]
  \scriptsize
  \centering
  \caption{Silhouette and orientation accuracy vs \# of model images}
    \begin{tabular}{cccc}
    \toprule
    \multicolumn{1}{P{7em}}{\# Model Images} & \multicolumn{1}{P{4.785em}}{Unoccluded  (IoU)} & \multicolumn{1}{P{3.785em}}{Occluded (IoU)} & \multicolumn{1}{P{8em}}{Mean Angle Error (Degrees)} \\
    \midrule
    \multicolumn{1}{P{7em}}{0 (class output)}  & 78.85 & 77.15 & 29.90 \\
    \multicolumn{1}{P{7em}}{0 (shared output)} & 77.87 & 74.95 & 31.32 \\
    1     & 89.20 & 86.31 & 14.27 \\
    4     & 89.38 & 86.05 & 13.60 \\
    6     & 89.54 & 86.36 & 15.19 \\
    12    & 88.68 & 85.25 & 13.48 \\
    \bottomrule
    \end{tabular}%
  \label{tab:model_input}%
\end{table}%




\section{Conclusion}
\label{sec:conclusion}

In this paper, we presented a method for object 6D pose estimation from monocular camera images, where detected object \ac{ROI} proposals are provided as input. We show that this method outperforms the state-of-the-art PoseCNN network and another recent RGB based method across the majority of object classes in the YCB-video dataset. The most significant contribution of our method is an intermediate silhouette representation for object viewpoints, which is shown to be a robust and effective abstraction from which to predict 3D orientation and also greatly reduces the sim-to-real domain shift when learning a model on synthetic data. This silhouette abstraction is demonstrated to improve accuracy of orientation predictions over previous methods. Also, by using an intermediate silhouette representation for detected objects, our method enables determining which parts of an object model are unoccluded in the scene. We proposed a novel strategy for predicting 3D translation from \ac{ROI} proposals, which does not suffer from ambiguities in apparent viewpoint, leading to improved translation accuracy over previous methods. Currently, SilhoNet predicts 3D orientations that are unique to symmetries in silhouette space. Future work will focus on extending this method to orientation predictions that are also unique in feature space, despite symmetries in object shape.

\renewcommand{\bibfont}{\normalfont\small}
\printbibliography

\end{document}